\documentclass[runningheads,a4paper]{llncs}

\usepackage{tikz}
\usepackage{caption}
\usetikzlibrary{arrows,positioning,shapes.geometric}
\usepackage{schemabloc}
\usetikzlibrary{circuits}
\usepackage{amssymb}
\usepackage{graphicx}
\usepackage{amsmath}

\usepackage{stackengine}
\usepackage[labelformat=empty,labelsep=quad,skip=3pt]{caption}
\usepackage{url}
\urldef{\mailsa}\path|{farazi, behnke}@ais.uni-bonn.de, niloofarazizi37@gmail.com|    
\newcommand{\keywords}[1]{\par\addvspace\baselineskip
\noindent\keywordname\enspace\ignorespaces#1}
\graphicspath{ {./} }
\usepackage{placeins}
\usepackage{float}
\usepackage{eso-pic}

\AtBeginDocument{\AddToShipoutPictureFG*{\AtTextUpperLeft{\put(0,\LenToUnit{40pt}){\parbox{\textwidth}{\centering\bfseries
27th International Conference on Artificial Neural Networks (ICANN), Rhodes, Greece, 2018
}}}}}

\begin{document}

\mainmatter  

\title{Location Dependency in Video Prediction}

\titlerunning{Location Dependency in Video  Prediction}

\author{Niloofar Azizi\thanks{Both authors contributed equally to this work.}
\and Hafez Farazi\footnotemark[1] \and Sven Behnke}
\authorrunning{Niloofar Azizi, Hafez Farazi, and Sven Behnke}

\institute{Bonn University, Computer Science Department,\\
 Endenicher Allee 19a, 53115 Bonn, Germany\\
\mailsa\\
\url{}}
\toctitle{Location Dependent Video Ladder Network}
\tocauthor{Niloofar Azizi, Hafez Farazi, and Sven Behnke}
\maketitle
\label{abstract-sec}
\begin{abstract}
Deep convolutional neural networks are used to address many computer vision problems, including video prediction. The task of video prediction
requires analyzing the video frames, temporally and spatially, and constructing a model of how the environment evolves. 
Convolutional neural networks are spatially invariant, though, which prevents them from modeling location-dependent patterns. 
In this work, the authors propose location-biased convolutional layers to overcome this limitation.
The effectiveness of location bias is evaluated on two architectures: Video Ladder Network (VLN) and Convolutional Predictive Gating Pyramid (Conv-PGP).
The results indicate that encoding location-dependent features is crucial for the task of video prediction. Our proposed methods significantly outperform spatially invariant 
models.
\end{abstract}
\vspace{-33px}
\label{Introduction-sec}
\section{Introduction}
\vspace{-9px}
\par The task of video prediction consists of predicting a set of successor 
frames, given a sequence of video frames. It is challenging, because the 
predictor needs to understand both contents and motion of the scene in order to 
make good predictions. 
In recent years, deep learning approaches became popular 
for video prediction. They analyze the video both spatially and temporally and 
learn hierarchical representations, which model the image evolution in terms of 
its content and dynamics~(\cite{mathieu2015deep}, \cite{wagnerlearning}). The learned representations can 
be used for a variety of applications, including action recognition and 
anticipating future actions, which can be utilized for instance in human-robot 
interaction scenarios. \par Convolutional deep learning architectures 
cannot recognize location-dependent features, however, due to the location-invariant nature of convolutions. In the task of the video prediction, for instance, learning the location of static obstacles in the environment leads to better frame forecasting. In 
this work, the authors propose three different methods to overcome this limitation:
\begin{itemize}
 \item[a)] encoding location features in separate channels of the input,
 \item[b)] convolutional layers with learnable location-dependent biases, and
 \item[c)] convolutional layers with learnable location-dependent biases and predefined location encodings.
\end{itemize}
\clearpage
\vspace{-15px}
\begin{figure}[t!]
{\hspace{12px}\includegraphics[width=0.90\textwidth]{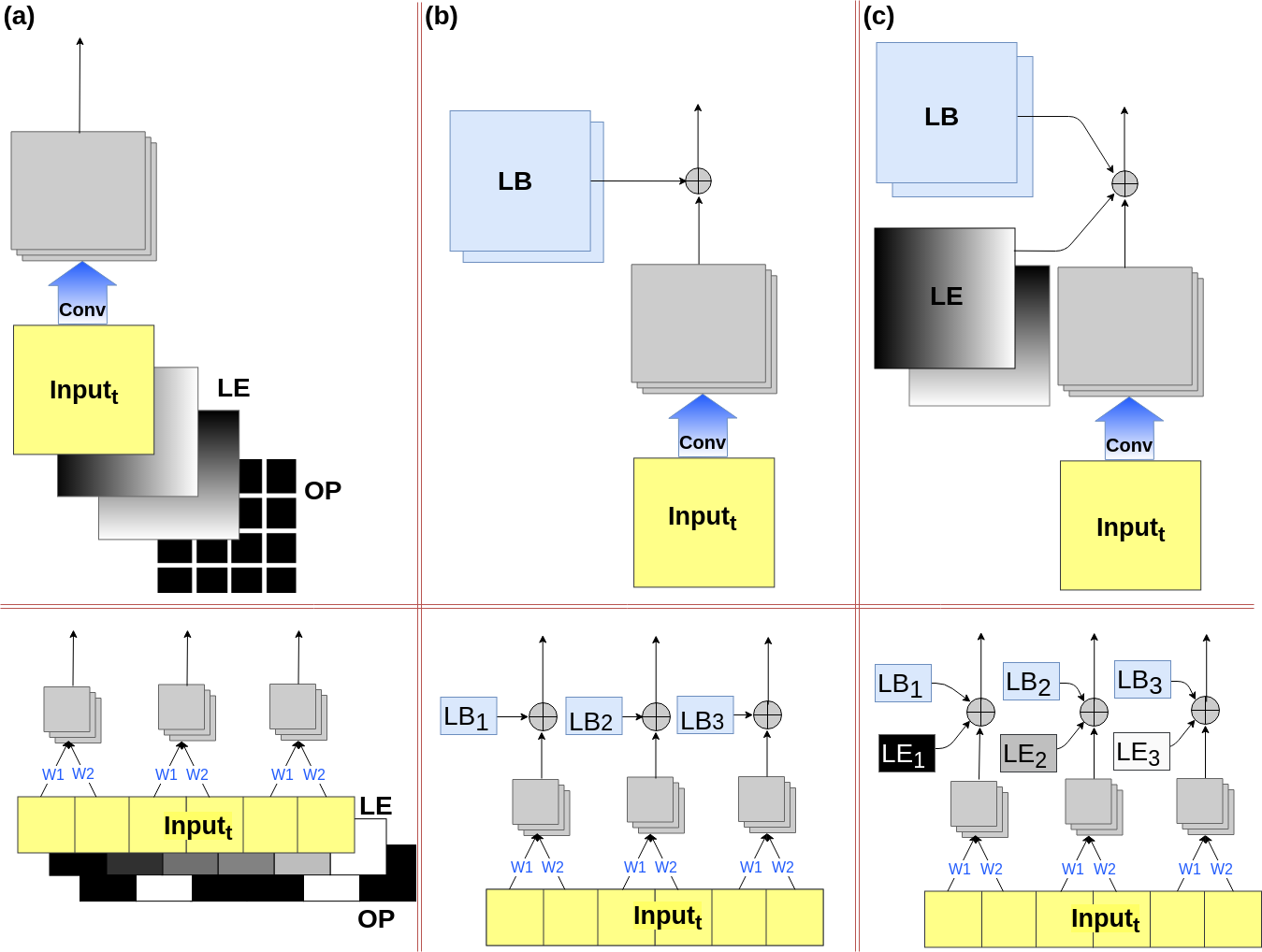}}   
\vspace{3px}
\caption{Proposed methods for location-dependency. Top row 2D and bottom row 1D.
a) three additional input channels, two of which encode location by gradients in x and y directions (LE); the third contains the occlusion pattern (OP). b) learnable location-dependent biases (LB) are added to the output of convolutions. c) learnable location-dependent biases and predefined location encodings use combined.}
\label{Teaser}
\end{figure}
\vspace{-13px}
These methods are illustrated in Fig.~\ref{Teaser} for 1D and two-dimensional convolutions.
\par We demonstrate the utility of our approach using two datasets that contain location dependencies.
The code and datasets of this paper are publicly available.\footnote{\url{https://github.com/AIS-Bonn/LocDepVideoPrediction}}
\label{Related_Work-sec}
\vspace{-13px}
\section{Related Work}
\vspace{-6px}
\par Convolutional deep learning architectures are spatially invariant, which leads 
to the constraint of not being able to model location-dependent patterns. \par 
To address this issue in various computer vision tasks, different approaches 
have been explored. Utilizing fully connected layers leads to learning 
location-dependent features, but this has the drawbacks of  
many parameters and no spatial weight sharing. In the PixelCNN architecture for conditional image generation, 
Oord et al.~\cite{van2016conditional} applied 1$\times$1 convolutions to map a hidden representation into a spatial representation. The disadvantage of this approach is that 
to extract the spatial features, a very large number of parameters is needed. In saliency prediction, Kruthiventi et 
al.~\cite{DeepFix} proposed adding another set of convolutional weights with the 
same size of the original filters. They convolved these additional weights with 
predefined fixed channels that encode the image center using Gaussian blobs with 
different horizontal and vertical extent. Ghafoorian et 
al.~\cite{ghafoorian2017location} applied specific location features to train 
the model and utilized location dependency for the task of brain 
MRI image segmentation. They showed that the results improve in comparison to CNNs that do 
not use location information. The above approaches depend all on predefined location feature structures. \par For the task of video prediction, different 
approaches have been explored. The most successful ones utilize deep learning 
methods. Cricri et al.~\cite{VLN} proposed Video Ladder Networks (VLN) by adding 
recurrent connections to the ladder network~\cite{rasmus2015semi}. Similar to 
ladder networks, VLN employs shortcut connections from the encoder to the 
respective decoder part, whereby it relieves the deeper layers from modeling 
details. The VLN architecture achieves a result competitive to VPN \cite{VPN} which is the state-of-the-art on the
synthetic dataset of Moving MNIST. However, the VLN architecture due to its convolutional 
layers, cannot deal with location-dependent features. Another recurrent network 
for the task of video prediction was proposed by Michalski et 
al.~\cite{michalski2014modeling}. Their PGP network is based on a 
gated autoencoder and a bilinear transformation model, to learn transformations 
between pairs of consecutive images 
(\cite{memisevic2013learning},~\cite{memisevic2010learning}). PGP is fully connected, which results in a large 
number of parameters. Its convolutional variant Conv-PGP reduces the number of parameters 
significantly~\cite{demodeling}, but looses the ability to learn location-dependent features.
For the evaluation of Conv-PGP, the authors augmented one-pixel padding to the 
input to learn a bouncing ball motion in their synthetic dataset.
\par While VLN 
and Conv-PGP have shown impressive performance in the task of video prediction, 
the above analysis shows that the effect of location-dependent features on 
these two architectures requires further investigation.
\label{VLN-Sec}
\vspace{-12px}
\section{Location Dependency in VLN Model} 
\vspace{-8px}
\par The VLN model~\cite{VLN} is a neural network architecture that predicts future frames 
by encoding the temporal and spatial features of a video. Although it achieves a competitive 
result in comparison to the state-of-the-art on Moving MNIST, 
due to the location invariant property of convolution operation, it cannot learn location-dependent features present in the dataset. The network would become unreasonably huge if 
we wanted to utilize a fully connected layer to allow for learning location-dependent features. Using a fully connected layer would also violate 
the assumption of weight sharing in the VLN architecture. The same-padding property around the border, which is not analyzed in the 
original paper, is the reason which allows 
the network to learn where to mirror digit velocity despite using only 
convolutional operations. Such a behavior is accidental, though, and should not be treated as a feature.

\par To demonstrate this limitation of the VLN architecture, we modified the Moving 
MNIST dataset to Occluded Moving MNIST, similar to what is used by Pr\'{e}mont-Schwarz et al.~\cite{RLN}. 
 As demonstrated in the experiment section, we tested the original one-layer VLN with this dataset and it did 
not achieve an acceptable result. 

\par To solve this issue, we propose three methods for providing location information to the network. 
In the first method illustrated in Fig.~\ref{Teaser}(a), we provide three additional input channels to the network: two gradient 
channels in $x$ and $y$ direction, starting from $0$ and ending with $1$, as well as one channel containing the occlusion 
grid pattern. The occlusion channel is 1 in the occlusion areas and 0 elsewhere. These additional input channels allow the network to 
infer the location-dependent feature of the border and to utilize the occlusion pattern.
In contrast to encoding location features in the original input channel, having 
additional channels does not alter the original input. Encooding occlusions in a separate channel can be useful, for example, when they are inferred from 
modalities other than a camera, like a laser scanner.
\par
\begin{figure}[!t]
\centering
 \hspace{-5px}
  \vspace{2px}
    \begin{minipage}[c]{0.55\textwidth}
\includegraphics[scale=0.28]{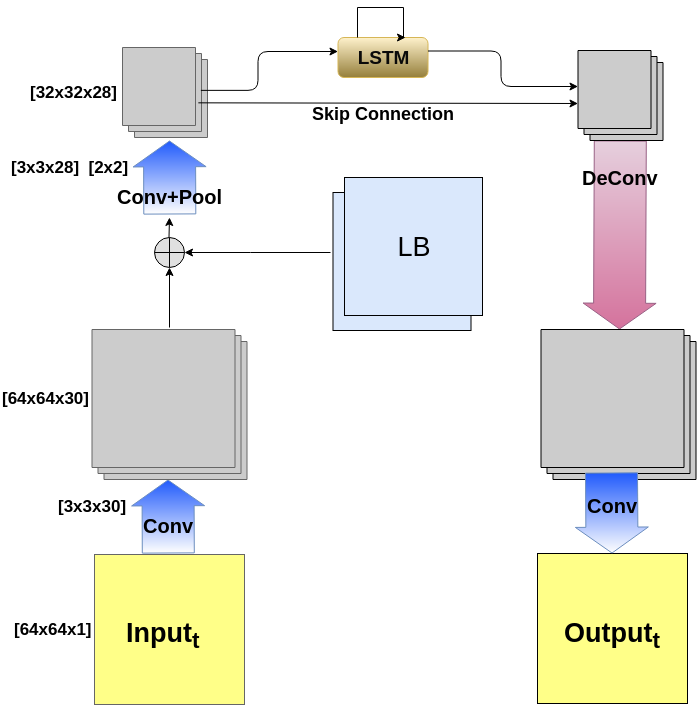}
  \end{minipage}\hfill
  \begin{minipage}[c]{0.38\textwidth}
    \vspace{90px}
\captionof{figure}{One-layer location-dependent VLN architecture consisting of two convolution layers in the encoder block, a Conv-LSTM block, and one deconvolution followed by a convolution layer in the decoder part. The trainable location-dependent bias (LB) is applied after the first convolution layer.}
\centering
\label{VLN-Model}
  \end{minipage}
\vspace{-16px}
\end{figure}

 In the second method~(Fig.~\ref{Teaser}(b) and Fig.~\ref{VLN-Model}), we replace the 
first convolutional layer in the encoder block with a location-dependent 
convolutional layer:
\vspace{-7.5px}
\begin{align}
LC(x,y) = A\bigg(\sum_{i,j}\Big(I(x+i, y+j)*W(i,j)+ b\Big) + W_1^{'}(x,y)+ W_2^{'}(x,y)\bigg)
\end{align}
\vspace{-13px}
\par where $A$ is the activation function. $W$ and $b$ are the weight and bias 
of the specified layer, respectively. Note that $b$ can be omitted, but we kept it to make the proposed layer easy to implement on top of an existing convolution layer. $I(x,y)$ is the input vector at 
the Cartesian position $(x,y)$ and $*$ represents the convolution operator. Note 
that $W_1^{'}$ and $W_2^{'}$ are location-dependent weights that are learned 
through the training procedure. $W_1^{'}$ and $W_2^{'}$ 
are shared for all convolutional filters, which is done by broadcasting over channel dimension.
\par In the third method, illustrated in Fig.~\ref{Teaser}(c), we added location-dependent gradients to the $W_1^{'}$ and $W_2^{'}$:
\vspace{-5px}
\begin{align}
 \vspace{-32px}
 \begin{split}
  \hspace{-5px}
\scriptstyle LC(x,y) = A\Big(\sum_{i,j}\big(I(x+i, y+j)*W(i,j)+ b\big)+\big(L_x(x, y) + W_1^{'}(x,y)\big)+\big(L_y(x, y) + W_2^{'}(x,y)\big)\Big)
\end{split}
\end{align}
\vspace{-7px}
\par where similar to additional input channels, $L_x(x, y)$ and $L_y(x, y)$ encode location by gradients in $x$ and $y$ directions, respectively.
Providing these facilitates the learning of more complex location-dependent biases. 
\vspace{-10px}
\label{Conv-PGP-Sec}
\section{Location Dependency in Conv-PGP Model}

\par PGP~\cite{michalski2014modeling} is designed based on the assumption that two temporally consecutive frames can be described as a linear transformation of each other. In the PGP architecture, by using a Gated AutoEncoder (GAE) as bi-linear model, the hidden
layer of mapping units $m$ encodes the transformation.
\vspace{3px}
\par The fully connected PGP architecture contains a significant number of parameters. To deal with this issue, we utilized its convolutional variant (Conv-PGP), similar to~\cite{demodeling},
where fully connected layers are replaced by convolutions. 
\par While Conv-PGP reduces the number of parameters significantly, it cannot learn location-dependent features such as the image border anymore.
Using valid convolutions prevents, e.g., learning the mirroring motion in the Bouncing Ball dataset. As 
shown in the experiment section, in the Conv-PGP model, the balls disappear instead of being reflected at the 
border which indicates that the model is incapable of predicting location-dependent motions.
\vspace{0.6px}
\par To demonstrate this limitation more clearly, we modified the Bouncing
Ball dataset. In the Occluded Bouncing Ball dataset, we augmented fixed strides of 
three pixels to occlude the moving balls as well as invisible lines to mirror 
the velocity. As shown in the following section, we trained the Conv-PGP with this dataset, and it did 
not achieve a satisfactory result.
To resolve this issue, we applied the three proposed methods for modeling location dependency to Conv-PGP. 
\vspace{-2px}
\label{Experiment}
\vspace{-8px}
\section{Experiment}
\vspace{-4px}
We tested our modified VLN architectures on the Occluded Moving MNIST dataset. Each video in the Occluded Moving MNIST dataset 
contains $10$ frames, with one MNIST digit moving inside a \mbox{64$\times$64} patch.
Digits are chosen randomly from the training set and
placed initially at random locations inside the patch with a random velocity. The frames are filled with occluding vertical and horizontal bars; the distance between them is eight pixels. 
In addition to that, we added invisible lines to mirror 
the velocity at a distance of ten pixels from the border. 

\par
\begin{figure}[!t]
 \vspace{-15px}
\begin{minipage}[c]{0.5\textwidth}
 \hspace*{30px}
\centering
\begin{tikzpicture}
\node[inner sep=0pt] (detection_result) at (0,0)
{\hspace{-35px}\includegraphics[scale=0.2]{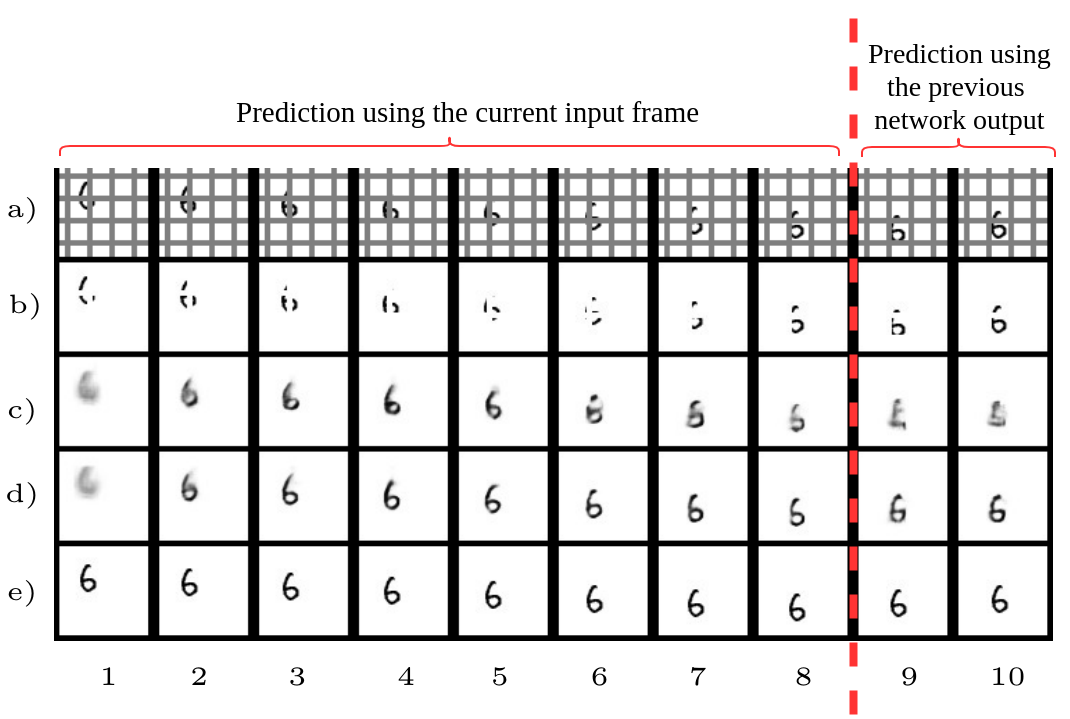}};
 \vspace*{-75px}     
\hspace{-45px}
\end{tikzpicture}
\end{minipage}\hfill
  \begin{minipage}[c]{0.35\textwidth}
  \vspace{20px}
\caption{Occluded Moving MNIST. a) Input frames with visualized occlusion. b) Frames given to the network. c) Predicted frames with the one-layer original VLN. d) Predicted frames with VLN-LDCAI. e) Expected ground truth frames.}
\label{VLN-Result}
 \end{minipage}
 \vspace{-23px}
\end{figure}

\begin{figure}[!b]
\begin{minipage}[c]{0.55\textwidth}
 \centering
     \stackunder{a)\hspace{5px}\includegraphics[width=0.95\textwidth]{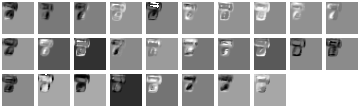}
     }{}
     \\
     \stackunder{b)\hspace{5px}\includegraphics[width=0.95\textwidth]{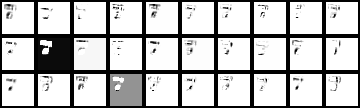}
     }{}
     \\
     \stackunder{c)\hspace{5px}\includegraphics[width=0.95\textwidth]{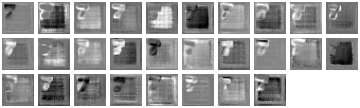}
     }{}
     \\
     \stackunder{d)\hspace{5px}\includegraphics[width=0.95\textwidth]{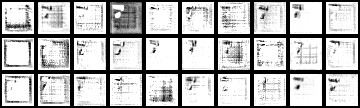}
     }{}
     \\
     \stackunder{e)\hspace{5px}\includegraphics[width=0.22\textwidth]{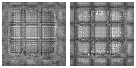}
     }
     {}
     \\
     \stackunder{f)\hspace{5px}\includegraphics[width=0.1\textwidth]{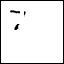}
     }{}
     \\
 \end{minipage}\hfill
  \begin{minipage}[c]{0.4\textwidth}
  \vspace{110px}
 \caption{Occluded Moving MNIST activities. a) Activation layers of Conv-LSTM block in original VLN. b) Activation layers of the encoder block in original VLN.
 Note that none of the channels can detect the location-dependent features. c) Activation layers of Conv-LSTM block in VLN-LDCAI. 
 d) Activation layers of the encoder block in VLN-LDCAI. e) Learned location-bias channels in VLN-LDCAI. f) Input frame. Note that the VLN-LDCAI automatically inferred location-dependent features.}
 \label{activation}
  \end{minipage}
 \end{figure}

 In our first experiment, we compare the one-layer original VLN architecture on Occluded Moving MNIST with our three proposed solutions:
\begin{itemize}
  \item VLN-AI: Two location gradient channels and one occlusion channel as additional location encoding inputs (Fig.~\ref{Teaser}(a)),
  \item VLN-LDC: Location-dependent bias in the encoder block (Fig.~\ref{Teaser}(b)), and
 \item VLN-LDCAI: Location-dependent bias in the encoder block and location gradient channels (Fig.~\ref{Teaser}(c)).
\end{itemize}
\vspace{-2px}
In our experiment, the first eight frames are predicted using the given frame from the dataset. The last two frames are predicted using the previous network output. 
Sample results of one-layer original VLN and VLN-LDCAI are depicted in 
Fig.~\ref{VLN-Result}. Sample activations of the Conv-LSTM and the encoder block for both the original VLN and the VLN-LDCAI are shown in 
Fig.~\ref{activation}.
These activations demonstrate that the original VLN cannot infer the 
location-dependent features while the VLN-LDCAI can learn location-dependent 
features including the border and the occlusion grid. 
\vspace{-5px}
\par Table ~\ref{VLNfinal_table_results} reports the prediction loss and the number of parameters for the evaluated model variant. It can be observed that all methods to model location dependencies improve performance.

\vspace{-23px}

\begin{table}[H]
\renewcommand{\arraystretch}{1.1}
\centering
\vspace{-10px}
\caption{Results of VLN models on Occluded Moving MNIST test dataset.}
\begin{tabular}
{l | c | c | c}
 Model\hspace{5px}& \hspace{1px}Prediction test loss (BCE)\hspace{5px} &\hspace{1px}Number of parameters\hspace{5px} \\
\hline \hline
VLN & 165.9   & 90K\\ 
VLN-AI  & 150.7   & 91K\\
VLN-LDC  & 154.7  & 103K\\
VLN-LDCAI \hspace{3px}& 153.2 & 103K\\
\end{tabular}
\vspace{5px}\\
\vspace{-23px}
\label{VLNfinal_table_results}
\end{table}
\par
\begin{figure}[H]
\vspace{-105px}
\centering
\includegraphics[width=1\textwidth]{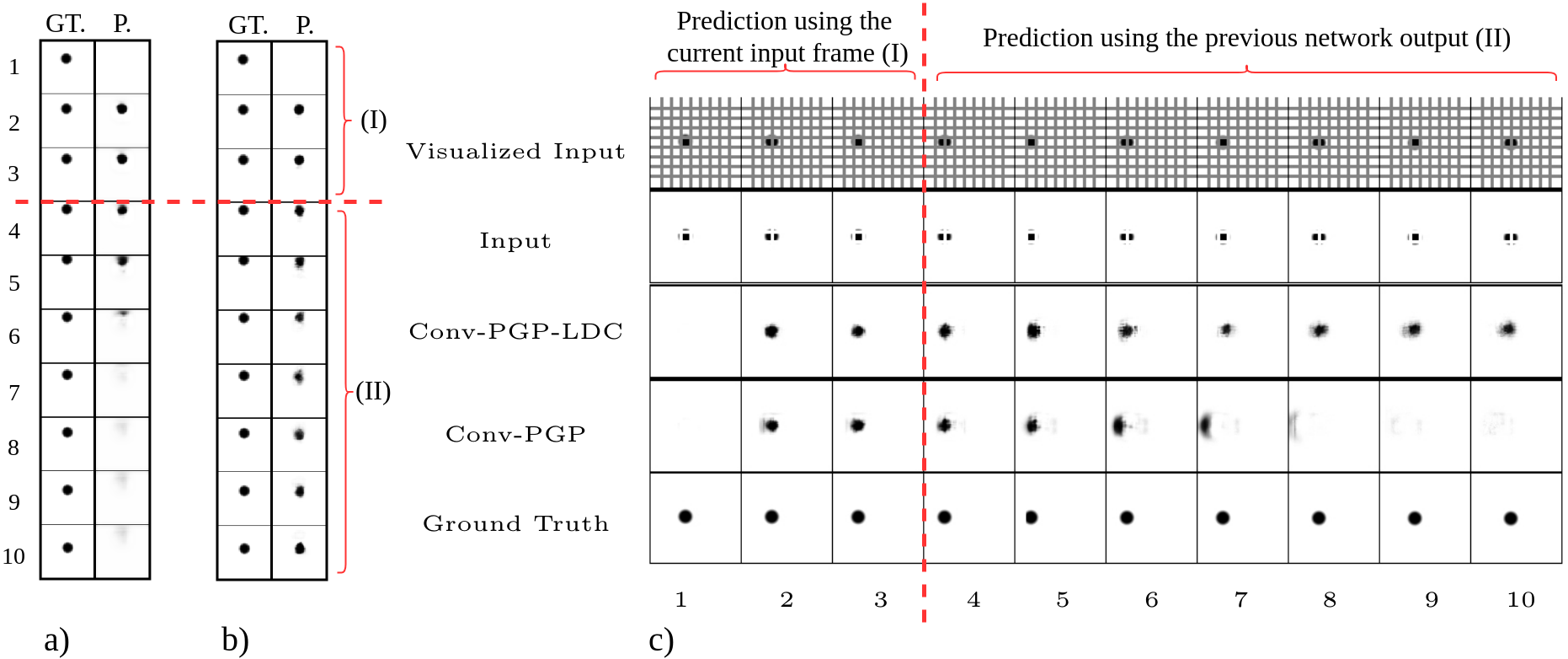}
\vspace{-10px}
\caption{Bouncing Ball results. a) Conv-PGP. b) Conv-PGP-AI. c) Conv-PGP-LDC and Conv-PGP on Occluded Bouncing Ball dataset.}
\label{pgpRes}
\end{figure}

 In a second experiment, we compared a one-layer Conv-PGP network with and without the border on the Occluded Bouncing Ball dataset, which is constructed similar to Occluded Moving MNIST. 
In our experiment, the first three frames are predicted using the given frame from the dataset. The last seven frames are predicted using the previous network output.
As illustrated in Figure~\ref{pgpRes}, learning the location-dependent 
features is crucial for the prediction task.
The prediction losses reported in Table ~\ref{PGPfinal_table_results} show that our proposed one-layer location-dependent Conv-PGP can solve the Occluded Bouncing Ball dataset and yields a much better result than one-layer Conv-PGP.

\vspace{90px}

\vspace{-31px}
\begin{table}[H]
\renewcommand{\arraystretch}{1.1}
\centering
\vspace{-10px}
\caption{Results of Conv-PGP models on Occluded Bouncing Ball test dataset.}
\begin{tabular}
{l | c | c | c}
 Model\hspace{5px} & \hspace{2px}Prediction test loss (BCE)\hspace{5px} &\ Number of parameters \hspace{3px} \\
\hline \hline
Conv-PGP  &   266.9 & 39k\\ 
Conv-PGP-AI   & 148.7   & 40k\\
Conv-PGP-LDC  &   139.4 & 56k\\ 
Conv-PGP-LDCAI \hspace{3px} &   143.3 & 56k\\
\end{tabular}
\vspace{-1px}\\

\vspace{-2px}
\label{PGPfinal_table_results}

\end{table}

\vspace{-35px}
\section{Conclusion}
\vspace{-7px}
\par Our experiments indicate that location information is a necessity in convolutional architectures for video prediction tasks as,
 for example, dealing with occlusions in the environment is challenging. 
To test three proposed variants of learning location-dependent features, we utilized the Occluded Moving MNIST and Occluded Bouncing Ball datasets which mimic occlusions in the real world. 
The proposed location-dependent inputs and biases allow the VLN and Conv-PGP models to learn more complex location-dependent features than just mirroring velocity at the borders. 
In contrast to previous approaches, our proposed learnable location-dependent biases do not assume any predefined underlying feature structure.
Our proposed location-dependent convolution layers significantly improve on the results of both one-layer VLN and one-layer Conv-PGP architectures.
\par In future work, we will explore the proposed methods for general deep convolutional neural network architectures, and test the performance on real-world datasets. 
\vspace{-10px}
\section*{Acknowledgment}
\label{acknowledgment}
\vspace{-5px}
This work was funded by grant BE 2556/16-1 (Research Unit FOR 2535 Anticipating Human Behavior) of the German Research Foundation (DFG).
\vspace{-10px}
\FloatBarrier
\bibliographystyle{unsrt}
\bibliography{sample}

\end{document}